\documentclass[runningheads]{llncs}
\usepackage{graphicx}
\usepackage{tikz}
\usepackage{comment}
\usepackage{amsmath,amssymb} 
\usepackage{color}

\usepackage{amsmath,amsfonts,bm}

\def\eqref#1{equation~\ref{#1}}

\def\1{\bm{1}}

\newcommand{\test}{\mathcal{D_{\mathrm{test}}}}

\DeclareMathAlphabet{\mathsfit}{\encodingdefault}{\sfdefault}{m}{sl}
\SetMathAlphabet{\mathsfit}{bold}{\encodingdefault}{\sfdefault}{bx}{n}

\newcommand{\E}{\mathbb{E}}

\DeclareMathOperator*{\argmax}{arg\,max}
\DeclareMathOperator*{\argmin}{arg\,min}

\usepackage{url}
\usepackage{soul,color}
\usepackage{verbatim}
\usepackage{amsfonts}
\usepackage{graphicx, enumerate, hyperref, color, bm, bbm, xcolor}
\usepackage{algorithm, algorithmicx, algpseudocode}
\usepackage{enumitem}
\usepackage{multirow}
\usepackage[toc,page]{appendix}
\usepackage{chngcntr}
\usepackage{color}
\usepackage{booktabs}

\def\E{\mathrm{E}}

\begin{document}
\pagestyle{headings}
\mainmatter
\def\ECCVSubNumber{1153}  

\title{Consistency-based Semi-supervised Active Learning: Towards Minimizing Labeling Cost} 

\titlerunning{Consistency-based Semi-AL}
%
\author{Mingfei Gao\inst{1}\thanks{Work done while the author was an intern at Google; now at Salesforce Research. Email: \email{mgao@cs.umd.edu}} \and
Zizhao Zhang\inst{2} \and
Guo Yu\inst{3} \and Sercan \"{O}. Ar{\i}k\inst{2} \and \\ Larry S. Davis\inst{1} \and Tomas Pfister\inst{2}}
%

\authorrunning{M. Gao, Z. Zhang et al.}
%
\institute{$^1$University of Maryland \ $^2$Google Cloud AI \ $^3$University of Washington\\
}
\maketitle

\begin{abstract}
  Active learning (AL) combines data labeling and model training to minimize the labeling cost by prioritizing the selection of high value data that can best improve model performance. 
  In pool-based active learning, accessible unlabeled data are not used for model training in most conventional methods.
  Here, we propose to unify unlabeled sample selection and model training towards minimizing labeling cost, and make two contributions towards that end. 
  First, we exploit both labeled and unlabeled data using semi-supervised learning (SSL) to distill information from unlabeled data during the training stage. 
  Second, we propose a consistency-based sample selection metric that is coherent with the training objective such that the selected samples are effective at improving model performance. 
  We conduct extensive experiments on image classification tasks. 
  The experimental results on CIFAR-10, CIFAR-100 and ImageNet demonstrate the superior performance of our proposed method with limited labeled data, compared to the existing methods and the alternative AL and SSL combinations. 
  Additionally, we also study an important yet under-explored problem -- ``When can we start learning-based AL selection?". We propose a measure that is empirically correlated with the AL target loss and is potentially useful for determining the proper starting point of learning-based AL methods.

\keywords{Active Learning, Semi-supervised Learning, Consistency-based Sample Selection.}
\end{abstract}
\section{Introduction}
\label{sec: introduction}

Deep learning models are improved when trained with more labeled data~\cite{Goodfellow-et-al-2016}. 
A standard deep learning procedure involves constructing a large-scale labeled dataset and optimizing a model on it. Yet, in many real-world scenarios, large-scale labeled datasets can be very costly to acquire, especially when expert annotators are required, as in medical diagnosis. 
An ideal framework would integrate data labeling and model training to improve model performance with minimal amount of labeled data. 

Active learning (AL) \cite{balcan2009agnostic} assists the learning procedure by judicious selection of unlabeled samples for human labeling,
with the goal of maximizing the model performance with minimal labeling cost. 
We focus on practically-common pool-based AL, where an unlabeled data pool is given initially and the AL mechanism iteratively selects batches to label in conjunction with training. 

Learning-based AL methods select a batch of samples for labeling with guidance from the previously-trained model and then add these samples into the labeled dataset for the model training in the next cycle. Existing methods generally start with a randomly sampled labeled set. The size of the starting set affects learning-based AL performance -- when the start size is not sufficiently large, the models learned in subsequent AL cycles are highly-biased which results in poor selection, a phenomenon commonly known as the \textit{cold start problem}~\cite{konyushkova2017learning,houlsby2014cold}. When cold start issues arise, learning-based selection yields samples that lead to lower performance improvement than naive uniform sampling \cite{konyushkova2017learning}.

To improve the performance at early AL cycles when the amount of labeled data is limited, it is important to address cold-start and ensure high performance later on with low labeling cost. 
Along this line of research, one natural idea for pool-based AL is integration of abundant unlabeled data into learning using semi-supervised learning (SSL)~\cite{zhu2003combining,tomanek2009semi}. 
Recent advances in SSL ~\cite{berthelot2019mixmatch,verma2019interpolation,xie2019unsupervised,sohn2020fixmatch,sohn2020simple} has demonstrated the vast potential of utilizing unlabeled data for learning effective representations. 
Although ``semi-supervised AL" seems natural, only a small portion of AL literature has focused on it. 
Past works that use SSL for AL ~\cite{drugman2019active,rhee2017active,zhu2003combining,sener2017active} treated SSL and AL independently without considering their impact on each other. We on the other hand, hypothesize that a better AL selection criterion should be in coherence with the corresponding objectives of unlabeled data in SSL to select the most valuable samples. 
A primary reason is that SSL already results in embodiment of knowledge from unlabeled data in a meaningful way, thus AL selection should reflect the value of additionally collected labeled data on top of such embodied knowledge. 
Based on these motivations, we propose an AL framework that integrates SSL to AL and also a selection metric that is highly related to the training objective.

The proposed AL framework is based on an insight that has driven recent advances in SSL~\cite{berthelot2019mixmatch,verma2019interpolation,xie2019unsupervised} -- a model should be consistent in its decisions between a sample and its meaningfully distorted versions, obtained via appropriate data augmentation.
This motivates us to introduce an AL selection strategy: \emph{a sample along with its distorted variants that yields low consistency in predictions indicates that the SSL model may be incapable of distilling useful information from that unlabeled sample -- thus human labeling is needed.}

\noindent Overall, \textbf{our contributions} are summarized as follows:
\begin{enumerate}
    \item We propose to unify model training and sample selection with a semi-supervised AL framework. The proposed framework outperforms the previous AL methods and the baselines of straightforward SSL and AL combinations.
    \item We propose a simple yet effective selection metric based on sample consistency which implicitly balances sample uncertainty and diversity during selection.  
    With comprehensive analyses, we demonstrate the rationale behind the proposed consistency-based sampling.
    \item We propose a measure that is potentially useful for determining the proper start size to mitigate cold start problems in AL.
\end{enumerate}

\begin{figure}[t]
    \centering
    \includegraphics[width=1.0\linewidth]{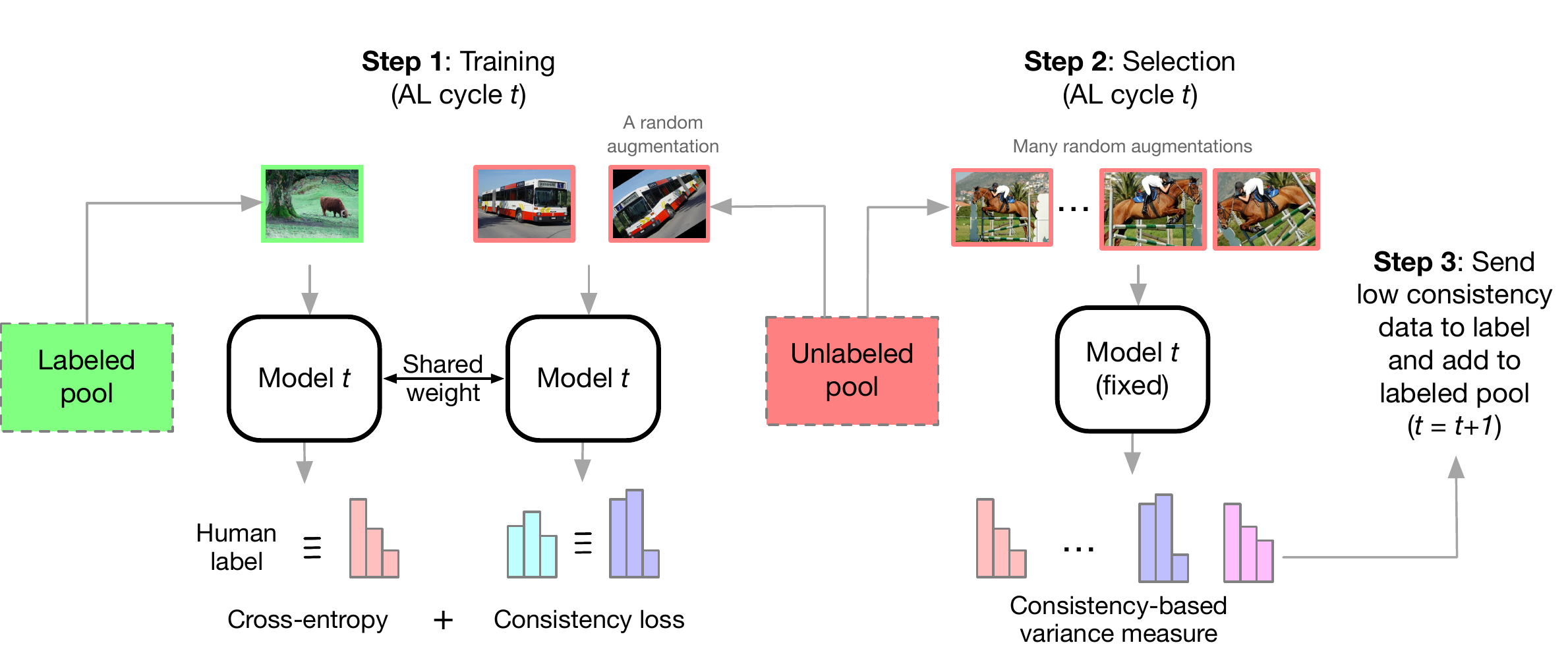}
    \caption{Illustration of the proposed framework at $t^{th}$ AL cycle. During training, both labeled and unlabeled data are used for the model optimization, with cross-entropy loss encouraging correct predictions for the labeled samples and consistency-based loss encouraging consistent outputs between unlabeled samples and their augmentations. 
    During sample selection, the unlabeled samples and their augmentations are evaluated by the model obtained from the training stage. 
    Their outputs are measured by our proposed consistency-based metric. 
    The samples with low consistency scores are selected for labeling and sent to the labeled pool}
    \label{fig: pipeline}
\end{figure}

\section{Related Work}
\subsection{Active learning}
There exists a broad literature on AL~\cite{dasgupta2008general,dasgupta2008hierarchical,balcan2009agnostic,Cortes2019graph}. Most AL methods can be classified under three categories: uncertainty-based methods, diversity-based methods and methods based on model performance change. Most uncertainty-based methods use \emph{max entropy}~\cite{lewis1994heterogeneous,lewis1994sequential} and \emph{max margin}~\cite{roth2006margin,balcan2007margin,joshi2009multi} criteria due to their simplicity.
Some others use distances between samples and the decision boundary~\cite{tong2001support,brinker2003incorporating}. Most uncertainty-based methods use heuristics, while recent work~\cite{yoo2019learning} directly learns the target loss of inputs jointly with the training phase and shows promising results. Diversity-based methods select diverse samples that span the input space maximally~\cite{nguyen2004active,mac2014hierarchical,hasan2015context,sener2017active}.
There are also methods that consider uncertainty and diversity in conjunction~\cite{guo2010active,elhamifar2013convex,yang2015multi}. The third category estimates the future model status and selects samples that encourage optimal model improvement~\cite{roy2001toward,settles2008multiple,freytag2014selecting}.

\subsection{Semi-supervised active learning}
Both AL and SSL aim to improve learning with limited labeled data, thus they are naturally related. Yet, only a few recent works have considered combining AL and SSL.
In \cite{drugman2019active}, joint application of SSL and AL is considered for speech understanding, and significant error reduction is demonstrated with limited labeled speech data. Their AL selection criteria is based on a confidence score obtained from the posterior probabilities of the decoded text. Rhee~\emph{et al.}~\cite{rhee2017active} propose an semi-supervised AL system which demonstrates superior performance in the pedestrian detection task. Zhu~\emph{et al.}~\cite{zhu2003combining} combine AL and SSL using Gaussian fields. Sener~\emph{et al.}~\cite{sener2017active} also consider SSL during AL cycles. However, in their setting, the performance improvement is marginal when adding SSL in comparison to their supervised counterpart, potentially due to the sub-optimal SSL method and combination strategy. Recently, Sinha~\emph{et al.} propose VAAL in~\cite{sinha2019variational}, where a variational autoencoder and an adversarial network are learned using both labeled and unlabeled samples to infer the representativeness of unlabeled samples during the sampling process. Although, unlabeled data is not used for model training. The concurrent AL works~\cite{simeoni2019rethinking}\cite{song2019combining} also consider integrating SSL, but their selection procedures are independent from the model training. We demonstrate that our proposed method unifying AL selection with SSL training is superior than the straightforward-combination strategy.

\subsection{Agreement-based active learning}
Agreement-based methods, also referred as ``query-by-committee", base the selection on the opinions of a committee which consists of independent AL metrics or models~\cite{seung1992query,cohn1994improving,mccallumzy1998employing,iglesias2011combining,beluch2018power,cortes2019active}. 
Our method is related to agreement-based AL where samples are determined based on the conformity of different metrics or models. Specifically, our method selects samples that mostly disagree with the predictions of their augmentations.

\section{Consistency-based Semi-supervised AL}
\label{sec: consistency}
\begin{algorithm}[h]
\caption{A semi-supervised learning based AL framework}
\label{alg:al}
\begin{algorithmic}
\Require Unlabeled data pool $\mathcal{D}$, the total number of steps $T$, selected sample batch set $B$, AL batch size $K$, start size $K_0 \ll |\mathcal{D}|$
\State $B_0 \gets$ uniformly sampling from $\mathcal{D}$ with $|B_0| = K_0$
\State $U_0 \gets \mathcal{D} \backslash B_0$
\State $L_0 \gets \left\{ (x, \mathcal{J}(x)): x \in B_0 \right\}$, where $\mathcal{J}(x)$ stands for the assigned label of $x$.
\For {$t = 0, \ldots, T - 1$}
    \State (training) $M_{t} \gets \argmin_{M} \left\{\frac{1}{|L_{t}|} \sum_{(x, y) \in L_{t}}\mathcal{L}_l (x, y, M) + \frac{1}{|U_{t}|}\sum_{x \in U_{t}}\mathcal{L}_u (x, M) \right\}$
    \State (selection) $B_{t + 1} \gets \argmax_{B \subset U_{t} } \left\{ \mathcal{C}(B, M_{t}), \; s.t. \; |B| = K\right\}$
    \State (labeling)  $L_{t + 1} \gets L_t \cup \left\{ (x, \mathcal{J}(x)): x \in B_{t + 1} \right\}$
    \State (pool update) $U_{t + 1} \gets U_t \setminus B_{t + 1}$
\EndFor
\State $M_{T} \gets \argmin_{M} \left\{\frac{1}{|L_{T}|} \sum_{(x, y) \in L_{T}}\mathcal{L}_l (x, y, M) + \frac{1}{|U_{T}|}\sum_{x \in U_{T}}\mathcal{L}_u (x, M) \right\}$
\State \Return $M_T$
\end{algorithmic}
\end{algorithm}

We consider the setting of pool-based AL, where an unlabeled data pool is available for selection of samples to label. 
To minimize the labeling cost, we propose a method that unifies selection and model updates. The proposed semi-supervised AL is depicted in Fig.~\ref{fig: pipeline}.

Most conventional AL methods base model learning only on the available labeled data, ignoring the useful information in the unlabeled data. While, we incorporate a semi-supervised learning (SSL) objective at training phases of AL cycles. The target model $M_{t}$ at AL selection cycle $t$ is learned by minimizing an objective loss function of the form $\mathcal{L}_l + \mathcal{L}_u$, where $L_l$ and $L_u$ indicate supervised and unsupervised losses, respectively. 
$\mathcal{L}_l$ is the supervised learning objective, such as the standard cross-entropy loss for classification. The proposed semi-supervised AL framework is presented in Algorithm \ref{alg:al}. For $\mathcal{L}_u$, we adopt the recent successful advances in SSL~\cite{athiwaratkun2018there,berthelot2019mixmatch,zhang2020distilling,verma2019interpolation}, that are based on minimizing the notion of sensitivity to perturbations with the idea of inducing ``consistency", i.e., imposing similarity in predictions when the input is perturbed in a way that would not change its perceptual content. 
For consistency-based SSL, the common choice for the loss is
\begin{equation}
 \mathcal{L}_u (x, M) = D(P(\hat{Y} = \ell|x, M), P(\hat{Y} = \ell|\tilde{x}, M)), 
\end{equation}
where $D$ is a distance function such as KL divergence~\cite{xie2019unsupervised}, or L2 norm \cite{laine2016temporal,berthelot2019mixmatch}, $M$ indicates the model and $\tilde{x}$ denotes a distortion (augmentation) of the input $x$.

The design of the selection criteria is crucial while integrating SSL into AL.
The unsupervised objective exploits unlabeled data by encouraging consistent predictions across slightly-distorted versions of each unlabeled sample. 
We hypothesize that \textit{labeling samples that have highly-inconsistent predictions should be valuable, because these samples are hard to be minimized using $\mathcal{L}_u$}. Human annotations on them can ensure a correct label, to be useful for supervised model training at next cycle. The samples that yield the large performance gains with SSL would not be necessarily the samples with the highest uncertainty, as the most uncertain data could be out-of-distribution examples, and including them in training might be misleading. Based on the intuitions, we argue that, for semi-supervised AL, valuable samples are the ones that demonstrate highly unstable predictions given different input distortions, i.e., the samples that a model can not consistently classify as a certain class. 

To this end, we propose a simple metric to quantify the inconsistency of predictions over a random set of data augmentations given a sample:
\begin{equation}
  \mathcal{E}(x, M) = \sum_{\ell = 1}^J \text{Var}\left[P(\hat{Y} = \ell|x, M), \, P(\hat{Y} = \ell|\tilde{x}_1, M), \, ..., \, P(\hat{Y} = \ell|\tilde{x}_N, M)\right],
  \label{eq:var}
\end{equation}
where $J$ is the number of response classes and $N$ is the number of perturbed samples of the original input data $x$, $\{\tilde{x}_1, ..., \tilde{x}_N\}$, which can be obtained by standard augmentation operations.

For batch selection, we jointly consider $K$ samples and aim to choose the subset $B$ such that the aggregate metric $\mathcal{C}(B, M) = \sum_{x \in B} \mathcal{E}(x, M)$ is maximized, i.e. the high inconsistency samples can be selected to be labeled by humans.

\section{Experiments}

In this section, we present experimental results of our proposed method. First, we compare our method to naive AL and SSL combinations, to show the effectiveness of our consistency based selection when all the methods are trained in a semi-supervised way. Second, since most recent AL methods still use only labeled data to conduct model training, we compare our method to recent AL methods and show a large improvement, motivating future research for semi-supervised AL. Third, we present qualitative analyses on several important properties of the proposed consistency based sampling.

\subsection{Experimental setup}

\noindent\textbf{Datasets}.
We demonstrate the performance of our method on CIFAR-10, CIFAR-100~\cite{krizhevsky2009learning} and ImageNet~\cite{imagenet_cvpr09} datasets. CIFAR-10 and CIFAR-100 have $60$K images in total, of which $10$K images are for testing. CIFAR-10 consists of 10 classes and CIFAR-100 has 100 classes. ImageNet is a large-scale image dataset with 1.2M images from 1K classes.

\noindent\textbf{Implementation details}.
Different variants of SSL methods encourage consistency loss in different ways. In our implementation, we focus on the recently-proposed method, Mixmatch~\cite{berthelot2019mixmatch}, which proposes a specific loss term to encourage consistency of unlabeled data. For comparison with selection baselines on CIFAR-10 and CIFAR-100, we use Wide ResNet-28~\cite{oliver2018realistic} as the base model and keep the default hyper-parameters for different settings following~\cite{berthelot2019mixmatch}. In each cycle, the model is initialized with the model trained in the previous cycle. $50$ augmentations of each image are obtained by horizontally flipping and random cropping, but we observe that $5$ augmentations can produce comparable results. To perform a fair comparison, different selection baselines start from the same initial model. The initial set of labeled data is randomly sampled and is uniformly distributed over classes. When comparing with advanced supervised AL methods, we follow~\cite{sinha2019variational} for the settings of start size, AL batch size and backbone architecture (VGG16~\cite{simonyan2014very}). We adopt an advanced augmentation strategy, RandAugment~\cite{cubuk2019randaugment}, to perform augmentation of unlabeled samples on ImageNet.

\noindent\textbf{Selection baselines}. 
We consider three representative selection methods. \emph{Uniform} indicates random selection (no AL). \emph{Entropy} is widely considered as an uncertainty-based baseline in previous methods~\cite{sener2017active,yoo2019learning}. It selects uncertain samples that have maximum entropy of its predicted class probabilities. \emph{k-center}~\cite{sener2017active} selects representative samples by maximizing the distance between a selected sample and its nearest neighbor in the labeled pool. We use the features from the last fully connected layer of the target model to compute sample distances.

\begin{figure}[t]
    \centering
    \includegraphics[width=1.0\linewidth]{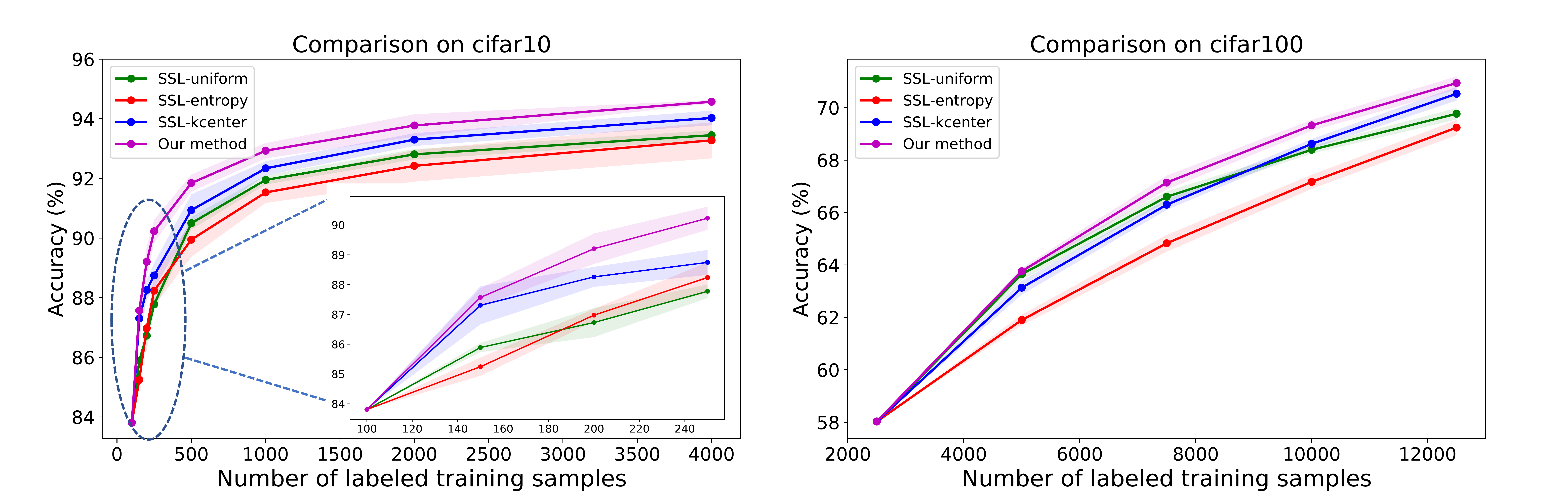}
    \caption{Model performance comparison with different sample selection methods on CIFAR-10 and CIFAR-100. Solid lines indicate the averaged results over 5 trials. Shadows represent standard deviation }
    \label{fig: comparison} 
\end{figure}
\begin{figure}[t]
    \centering
    \includegraphics[width=1.0\linewidth]{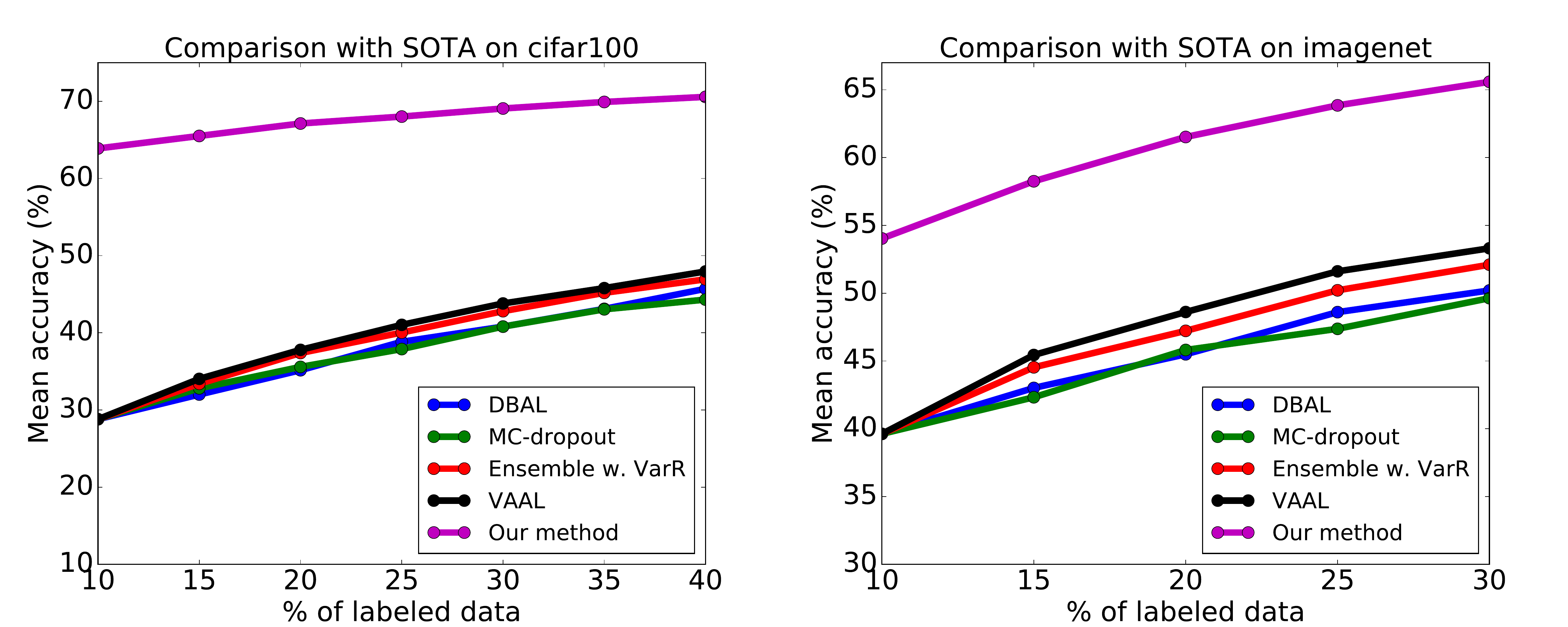}
    \caption{Comparison with recent AL methods on CIFAR-100 and ImageNet. Our results on CIFAR-100 and ImageNet are averaged over 5 and 3 trials, respectively}
    \label{fig: comp_soa}
\end{figure}
\begin{table}[t]
    \centering
    \caption{Comparison of different sampling methods on CIFAR-10. Note that all the methods are under the SSL setting and start with 100 labeled samples. When the number of labeled samples reaches $250$, AL batch size $K$ is set to be 250 and doubled afterwards. The reported results are averaged over 5 trials}
    \begin{tabular}{lccccc}
    \toprule
    \multirow{2}{*}{Methods}  & \multicolumn{4}{c}{ \# of labeled samples in total} \\ \cmidrule{2-6}
     & 250 & 500 & 1000 & 2000 & 4000 \\
    \midrule
    Uniform & 87.78$\pm$0.23 & 90.50$\pm$0.21 & 91.95$\pm$0.15 & 92.81$\pm$0.17 & 93.45$\pm$0.16 \\
    Entropy & 88.24$\pm$0.51 & 89.95$\pm$0.58 & 91.53$\pm$0.35 & 92.42$\pm$0.53 & 93.28$\pm$0.61 \\
    k-center & 88.75$\pm$0.42 & 90.94$\pm$0.53 & 92.34$\pm$0.24 & 93.30$\pm$0.21 & 94.03$\pm$0.25 \\
    Ours & \textbf{90.23$\pm$0.39} & \textbf{91.84$\pm$0.29} & \textbf{92.93$\pm$0.26} & \textbf{93.78$\pm$0.38} & \textbf{94.57$\pm$0.06} \\
    \bottomrule
    \end{tabular}
    \label{tab: comp_cifar10}
\end{table}

\subsection{Comparison with selection baselines under SSL}
\label{sec: comp_baselines}
To demonstrate the effectiveness of our method over the straightforward AL and SSL combinations, we focus on comparing with different selection methods in SSL framework. Fig.~\ref{fig: comparison} and Table~\ref{tab: comp_cifar10} show that when integrated with SSL training, our method outperforms baselines by a clear margin: on CIFAR-10, with $250$ labeled images, our method outperforms \emph{uniform} (passive selection) by $\sim2.5\%$ and outperforms~\emph{k-center}, by $\sim1.5\%$. As the number of labels increases, it is harder to improve model performance, but our method outperforms the \emph{uniform} selection with $4$K labels using only $2$K labels, halving the labeled data requirements for the similar performance.
Given access to all the labels ($50$K) for the entire training set, a fully-supervised model achieves an accuracy of $95.83\%$~\cite{berthelot2019mixmatch}. Our method with 4K (8\%) examples achieves about $30\%$ more error compared to the fully supervised method. 
CIFAR-100 is a more challenging dataset as it has the same amount of training images of CIFAR-10, but $10\times$ more categories. On CIFAR-100, we observe a consistent outperformance over baselines of our method at all AL cycles.

There is typically a trade-off between using a large and a small AL batch sizes. A large batch size will lead to insufficient usage of active learning given a limited budget. However, selecting a small batch of samples would lead to more AL cycles, which is computationally expensive. We conduct experiments on CIFAR-10 following the setting in Fig.~\ref{fig: comparison} using reasonable AL batch sizes. Results show that when consuming 200 labels in total, our methods obtain comparable performance (89.5\%, 89.2\% and 89.3\%) with AL batch size set to be 25, 50 and 100, respectively.

\subsection{Comparison with supervised AL methods}
We have shown that our method clearly outperforms the straightforward AL and SSL combinations in Sec.~\ref{sec: comp_baselines}. As mentioned, most AL methods focus on learning with only labeled samples. Consequently, it is worth showing the overall gap between our proposed framework and the existing methods to emphasize the benefit of the proposed framework. We choose the following recent methods as baselines: 
MC-Dropout~\cite{gal2016dropout}, DBAL~\cite{gal2017deep}, Ensembles w. VarR~\cite{beluch2018power} and VAAL~\cite{sinha2019variational} and compare with them on CIFAR-100 and ImageNet. The results of the baselines are reprinted from~\cite{sinha2019variational}.

As can be seen from Fig.~\ref{fig: comp_soa}, our method significantly outperforms the existing supervised AL methods at all AL cycles on both datasets. Specifically, when 40\% images are labeled, our method improves the best baseline (VAAL) by 22.62\% accuracy on CIFAR100 and by 12.28\% accuracy on ImageNet. The large improvements are mostly due to effective utilization of SSL at AL cycles.

Moreover, the performance of our method over the supervised models combined with the selection baselines in the scenario of very few labeled samples is of interest. As shown in Table~\ref{tab: SSL_SL}, our method significantly outperforms the methods which only learn from labeled data at each cycle. When $150$ samples in total are labeled, our method outperforms \emph{kcenter} by $39.24\%$ accuracy. 
\begin{table}[t]
\centering
\caption{Comparison between our method (trained in SSL) and our baselines that trained in supervised setting with very few labeled samples on CIFAR-10. All methods start from 100 labeled samples. The following columns are results of different methods with the same selection batch size. The reported results are over 5 trials}
\begin{tabular}{lccccc}
  \toprule
 \multirow{2}{*}{Setting} & \multirow{2}{*}{Methods}  & \multicolumn{4}{c}{ \# of labeled samples in total} \\\cmidrule{3-6}
      &  &100 & 150 & 200 & 250 \\
 \midrule
 \multirow{3}{*}{Supervised} 
 & Uniform &\multirow{3}{*}{41.85} & 46.13$\pm$0.38 & 51.10$\pm$0.60 & 53.45$\pm$0.71 \\
 & Entropy &  & 46.05$\pm$0.34 & 50.15$\pm$0.79 & 52.83$\pm$0.82 \\
                     & k-center & & 48.33$\pm$0.49 & 50.96$\pm$0.45 & 53.77$\pm$1.01 \\ \midrule
                    
 \multirow{1}{*}{Semi-supervised}
 & Ours &  \multirow{1}{*}{83.81}   & \textbf{87.57$\pm$0.31} & \textbf{89.20$\pm$0.51} & \textbf{90.23$\pm$0.49}\\
 \bottomrule
 \end{tabular}

 \label{tab: SSL_SL}
 \end{table}

\subsection{Analyses of consistency-based selection}
To build insights on the superior performance of our AL selection method, we analyze different attributes of the selected samples, which are considered to be important for AL. Experiments are conducted on CIFAR-10.
\begin{figure}[t]
    \centering
    \includegraphics[width=0.95\linewidth]{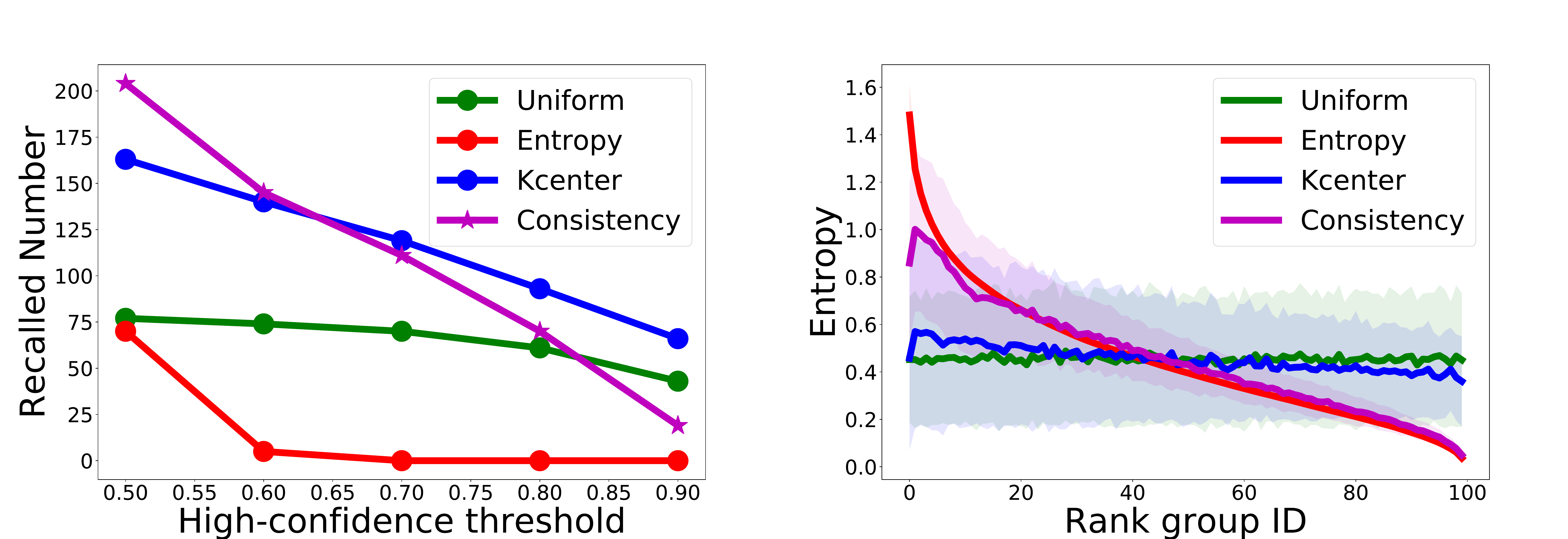}
    \caption{Left: Number of overconfident mis-classified samples in top $1\%$ samples ranked by different methods. Overconfident samples are defined as those having the highest class probability larger than threshold. Right: the average entropy of unlabeled samples ranked by different selection metrics. The ranked samples are divided into 100 groups for computing average entropy. Shadows represent standard deviation}
    \label{fig: uncertainty}
\end{figure}

\begin{figure}[t]
    \centering
    \includegraphics[width=0.95\linewidth]{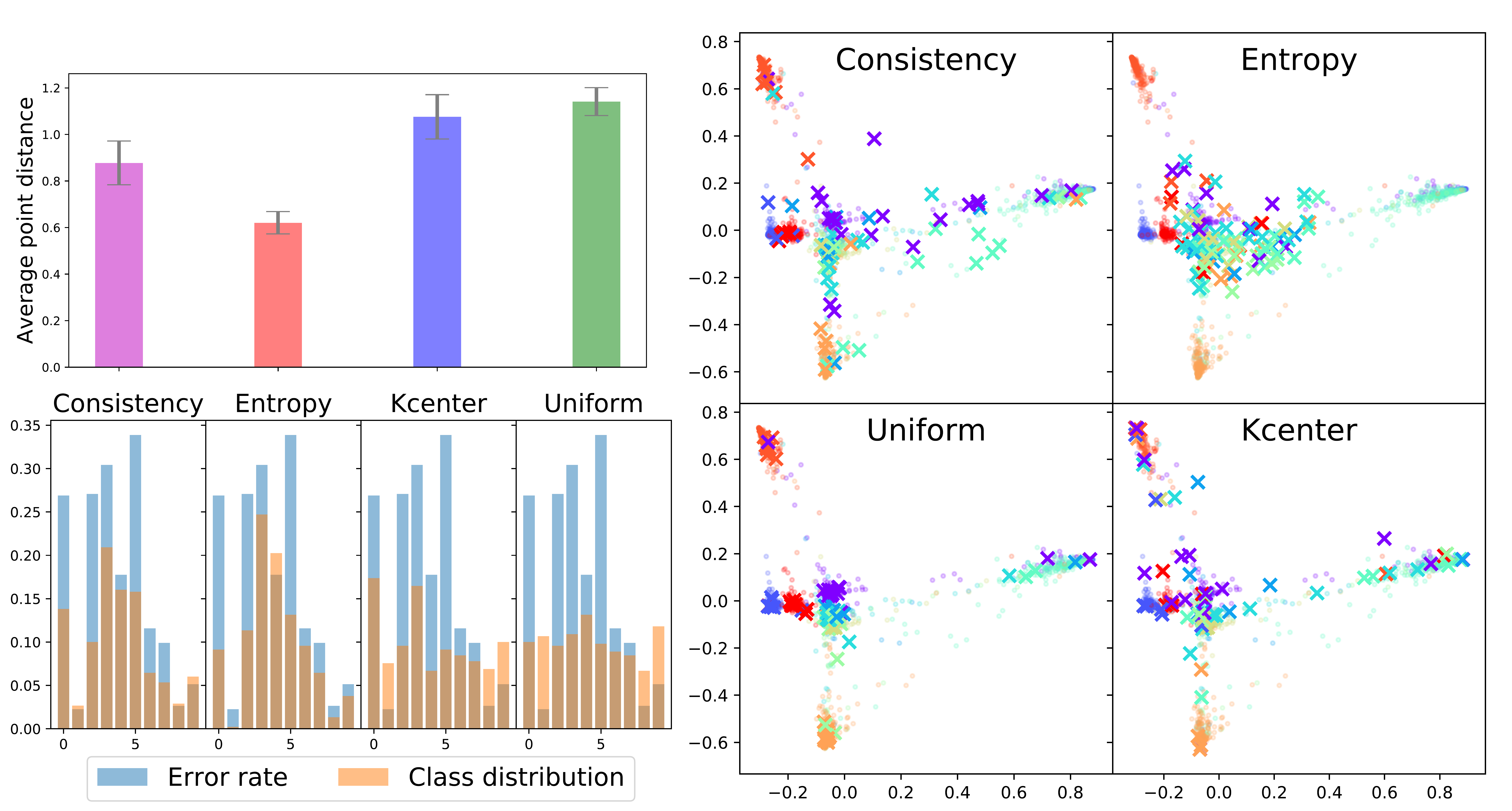}
    \caption{Average distance between samples (top-left): the average pair-wise $L_2$ distance of top $1\%$ unlabeled samples ranked by different selection metrics.
    Per-class error rate vs. the class distribution of the selected samples are shown in bottom-left. Diversity visualization (right): Dots and crosses indicate unlabeled (un-selected) samples and the selected samples (top 100), respectively. Each color represent a ground truth class
    }
    \label{fig: diversity}
\end{figure}

\noindent\textbf{Uncertainty and overconfident mis-classification}. 
Uncertainty-based AL methods query the data samples close to the decision boundary.
However, deep neural networks yield poorly-calibrated uncertainty estimates when the raw outputs are considered -- they tend to be overconfident even when they are wrong \cite{guo2017calibration,lakshminarayanan2017simple}. 
Entropy-based AL metrics would not distinguish such overconfident mis-classifications, thus may result in sub-optimal selection.
Fig.~\ref{fig: uncertainty} (left) demonstrates that our \emph{consistency}-based selection is superior in detecting high-confident mis-classification cases compared to \emph{entropy}-based selection. 
Fig.~\ref{fig: uncertainty}(right) shows the uncertainty in the selected samples with different methods, quantified using entropy. 
When different AL selection methods are compared, \emph{uniform} and \emph{k-center} methods do not base selection on uncertainty at all, whereas \emph{consistency} tends to select highly-uncertain samples but not necessarily the top ones. Such samples might contribute to our superior performance compared to \emph{entropy}. 

\noindent\textbf{Sample diversity}.
Diversity has been proposed as a key factor for AL \cite{yang2015multi}. 
\emph{k-center} is a diversity based AL method, preferring to select data points that span the whole input space.
Towards this end, Fig.~\ref{fig: diversity} (right) visualizes the diversity of samples selected by different methods. We use principal component analysis to reduce the dimensionality of embedded samples to a two-dimensional space. 
\emph{Uniform} chooses samples equally-likely from the unlabeled pool. 
Samples selected by \emph{entropy} are clustered in certain regions. 
On the other hand, \emph{consistency} selects data samples as diverse as those selected by \emph{k-center}. The average distances between top $1\%$ samples selected by different methods are shown in Fig.~\ref{fig: diversity} (top-left). We can see that \emph{entropy} chooses samples relatively close to each other, while \emph{consistency} yield samples that are separated with much larger distance which are comparable to samples selected by \emph{uniform} and \emph{k-center}.

\noindent\textbf{Class distribution complies with classification error}. 
Fig.~\ref{fig: diversity} (bottom-left) shows the per-class classification error and the class distribution of samples selected by different metrics.
Samples selected by \emph{entropy} and \emph{consistency} are correlated with per class classification error, unlike the samples selected by \emph{uniform} and \emph{k-center}.

\begin{figure}[htbp]
    \centering
    \includegraphics[width=0.95\linewidth]{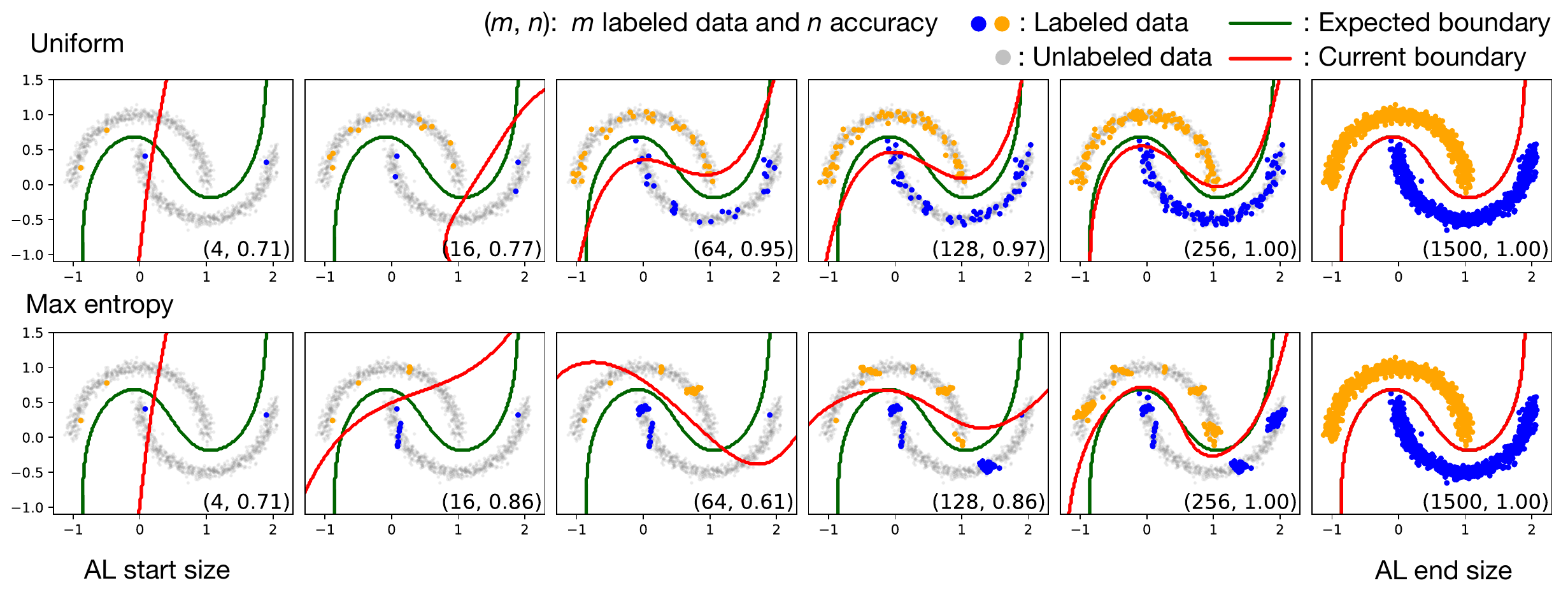}
    \caption{Illustration of cold-start problems for uncertainty-based AL.
    When the learned decision boundary is far away from the expected boundary (the boundary when all labels are available for the entire training set), e.g. the second and third columns, the selected samples by uncertainty-based AL is biased, leading to sub-optimal performance }
    \label{fig: moon}
\end{figure}

\section{When can we start learning-based AL selection?}
Based on the studies above, our proposed semi-supervised AL framework demonstrates clear advantages. While towards minimizing the labeling cost, a challenging issue, cold start failure, may occur when only a extreme small labeled set is available, which leads to sub-optimal AL performance. The proper study of this problem is essential for scenarios especially when labels are very expensive or challenging to collect.

\subsection{Cold-start failure}
When the size of the initial labeled dataset is too small, the learned decision boundaries could be far away from the real boundaries and AL selection based on the model outputs could be biased.
To illustrate the problem, Fig.~\ref{fig: moon} shows the toy two-moons dataset using a simple support vector machine model with an RBF kernel, trained in supervised setting to learn the decision boundary \cite{oliver2018realistic}. 
As can be seen, the naive uniform sampling approach achieves better predictive accuracy by exploring the whole space. On the other hand, the samples selected by \emph{max entropy} concentrate around the poorly-learned boundary. 

\begin{figure}[t]
    \centering
    \includegraphics[width=0.95\linewidth]{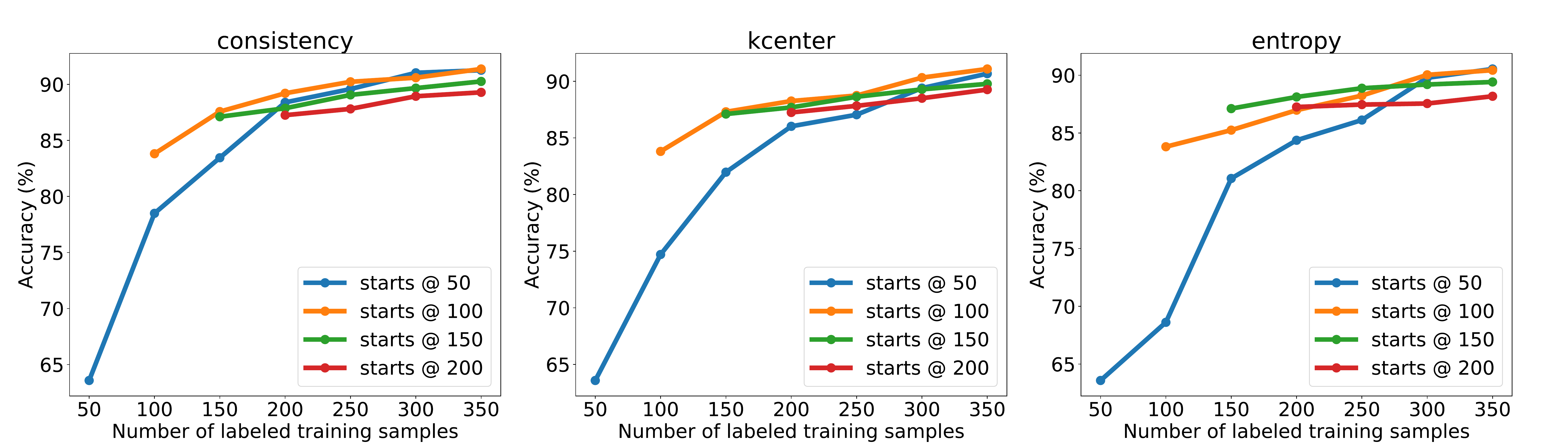}
    \caption{Comparison of different sampling methods trained with SSL framework on CIFAR-10 when AL starts from different number of labeled samples}
    \label{fig: starts_comp}
\end{figure}

Next, we study the cold start phenomenon for our proposed semi-supervised AL method. We focus on CIFAR-10 with small labeled initial sets, shown in Fig.~\ref{fig: starts_comp}.
Using uniform sampling to select different starting sizes, AL methods achieve different accuracies. 
For example, the model starting with $K_0 = 50$ data points clearly under-performs the model starting with $K_0 = 100$ samples, when both models reach 150 labeled samples. It may be due to the cold start problem encountered when $K_0 = 50$. On the other hand, given a limited labeling budget, naively choosing a large start size is also not practically desirable, because it may lead to under-utilization of learning-based selection. For example, starting with $K_0 = 100$ labeled samples has better performance than starting from 150 or 200, since we have more AL cycles in the former case given the same label budget. 
The semi-supervised nature of our learning proposal encourages the practice of initiating learning-based sample selection from a much smaller start size.
However, the initial model can still have poorly-learned boundary when started with extremely small labeled data. If there is a sufficiently large validation dataset, this problem can be relieved by tracking validation performance. However, in practice, such a validation set typically doesn't exist.
These motivate us to conduct an exploration to systematically infer a proper starting size.
\begin{figure}[t]
    \centering
    \includegraphics[width=0.95\linewidth]{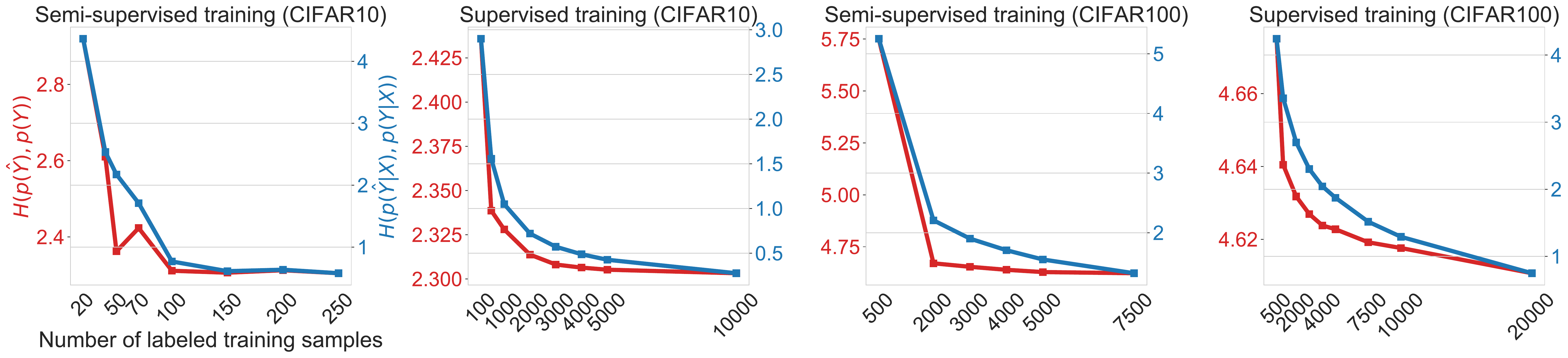}
    \caption{Empirical risk (i.e. the target loss) on the entire training data (in blue) and cross-entropy between $p(\hat{Y})$ and $p(Y)$ show strong correlations in both semi-supervised and supervised settings}
\label{fig: entropy}
\end{figure}

\subsection{An exploratory analysis in start size selection}
Recall from the last step of Algorithm \ref{alg:al}, if $T$ is set such that $U_T = \emptyset$, i.e., if the entire dataset is labeled, then the final model $M_T$ is trained to minimize the purely supervised loss $\mathcal{L}_l$ on the total labeled dataset $L_T$. Consider the cross-entropy loss for any classifier $p(\hat{Y} | X)$, which we call the \textit{AL target loss}:
\begin{equation}
   \mathcal{L}_l \left[L_T, p(\hat{Y} | X)\right] = - \frac{1}{|L_T|} \sum_{(x, y) \in L_T}  \log p(\hat{Y} = y| X = x).
   \label{eq:emp}
\end{equation}
Note that the goal of an AL method can be viewed as minimizing the AL target loss of the entire training set $L_T$ \cite{zhu2003combining} with the small subset of labeled data available. 
In any intermediate AL step, we expect our model can minimize the target loss.
If the model do a poor job in approximating and minimizing Eq.~\ref{eq:emp} (cold start problem occurs), the quality of the samples selected in the subsequent AL cycles could be consequently poor.
Therefore, it is crucial to assess the performance of the currently-learned model in minimizing the criterion in Eq.~\ref{eq:emp}. However, since the labeled data set $L_t$ at cycle $t$ is a strict subset of the entire training set $L_T$, it is impossible to simply plug the most recently-learned model $\hat{Y}$ in Eq.~\ref{eq:emp} for direct calculation.

To this end, we approximate the target loss based on the following proposition (see proof in the supplementary material), which gives upper and lower bounds on the expected loss: 
\begin{proposition} \label{prop:risk}
   For any given distribution of $Y$, and any learned model $\hat{Y}$, we have
    \begin{align}
   H \left[ p(Y) , p(\hat{Y}) \right] - H[p(X)] &\leq R_H\left[p(\hat{Y} | X)\right] = \E_{X} \left\{ H \left[ p(Y | X), p(\hat{Y} | X) \right] \right\}  \nonumber\\
   &\leq H \left[ p(Y) , p(\hat{Y}) \right] - H[p(X)] - \log \hat{Z},
    \label{eq:bounds}
    \end{align}
    where $H[p, q]$ is the cross-entropy between two distributions $p$ and $q$, $H[p(X)]$ is the entropy of the random variable $X$, and $\hat{Z} = \min_{x, y} p(X = x | \hat{Y} = y)$ .
\end{proposition}
Proposition \ref{prop:risk} indicates that the AL target loss, \emph{i.e.}, $R_H\left[p(\hat{Y} | X)\right]$, can be both upper and lower bounded. In particular, both bounds involve the quantity $H[p(Y), p(\hat{Y})]$, which suggests that $H[p(Y), p(\hat{Y})]$ could potentially be tracked to analyze $R_H[p(\hat{Y}| X)]$ when different numbers of samples are labeled. Unlike the unavailable target loss on the entire training set, $H[p(Y), p(\hat{Y})]$ does not need all data to be labeled. In fact, to compute $H[p(Y), p(\hat{Y})]$, we just need to specify a distribution for $Y$, which could be assumed from prior knowledge or estimated using all of the labels in the starting cycle.

As shown in Fig.~\ref{fig: entropy}, we observe a strong correlation between the target loss and  $H[p(Y), p(\hat{Y})]$, where $Y$ is assumed to be uniformly distributed. In practice, a practitioner can trace the difference of $H[p(Y), p(\hat{Y})]$ between two consecutive points and empirically stop expanding the start set when the difference is within a pre-defined threshold. Particularly, in SSL setting, $100$ or $150$ labeled samples may be used as start set on CIFAR-10, as the value of $H[p(Y), p(\hat{Y})]$ essentially ceases decreasing, which coincides with the oracle stopping points if we were given access to the target loss. In contrast, a start size of $50$ may not be favorable since the difference of $H[p(Y), p(\hat{Y})]$ between the points of $50$ and $20$ are relatively large. A similar pattern in the supervised learning setting is also shown in Fig.~\ref{fig: entropy}.
\begin{table}[t]
\centering
\caption{Comparison of different sampling methods in the supervised setting on CIFAR-10. All methods start from 1000 labeled samples. The reported results are over 5 trials}
\begin{tabular}{lccccc}
\toprule
\multirow{2}{*}{Methods}  & \multicolumn{5}{c}{ \# of labeled samples in total} \\ \cmidrule{2-6}
 & 1000 & 1500 & 2000 & 2500 & 3000 \\
\midrule
Uniform & \multirow{4}{*}{72.93} & 75.38$\pm$0.17 & 77.46$\pm$0.3 & 78.79$\pm$0.38 & 80.81$\pm$0.28 \\
Entropy & & 76.31$\pm$0.18 & 79.50$\pm$0.29 & 81.30$\pm$0.31 & 82.67$\pm$0.55 \\
k-center & & 74.25$\pm$0.29 & 77.56$\pm$0.30 & 79.50$\pm$0.20 & 81.70$\pm$0.32 \\
Ours & & 76.63$\pm$0.17 & 79.39$\pm$0.31 & 80.99$\pm$0.39 & 82.75$\pm$0.26  \\
\midrule
\end{tabular}
\label{tab: comp_cifar10_sup}
\end{table}

\section{Weaknesses of our method}
We explore how well our AL selection method would perform with supervised learning using only labeled samples. Following~\cite{yoo2019learning}, we start with 1000 labeled samples on CIFAR-10. As shown in Table~\ref{tab: comp_cifar10_sup}, after 4 AL cycles ($B=500$, totaling 3000 labels), \emph{uniform}, \emph{k-center}, \emph{entropy} and our method (\emph{consistency}) achieve accuracy of $80.81\%$, $81.70\%$, $82.67\%$ and $82.75\%$, respectively. It shows that \emph{consistency} sampling performs comparable with the baseline metrics without significant improvement. This discourages the direct application of our selection metric in the supervised setting. Mixmatch is mainly used as the target model in this work and we experiment with two more SSL methods (see results in the supplementary material). However, comprehensive analyses with extensive SSL methods are desirable to further understand the advantages/disadvantages of our approach. As an exploratory analysis, we propose a measure that is shown to be strongly correlated with the AL target loss, but exact determination of the optimal start size is yet to be addressed.

\section{Conclusion}
We present a consistency-based semi-supervised AL framework and a simple pool-based AL selection metric to select data for labeling by leveraging unsupervised information of unlabeled data during training. Our experiments demonstrate that our semi-supervised AL method outperforms the state-of-the art AL methods and also alternative SSL and AL combinations. Through quantitative and qualitative analyses, we show that our proposed metric implicitly balances uncertainty and diversity when making selection. 
In addition, we study and address the practically-valuable and fundamentally-challenging question -- ``When can we start learning-based AL selection?". We present a measure to assist determining proper start size. Our experimental analysis demonstrates that the proposed measure correlates well with the AL target loss (i.e. the ultimate supervised loss on all labeled data), thus is potentially useful to evaluate target models without extra labeling effort. Overall, semi-supervised AL opens new horizons for training with very limited labeling budget, and we highly encourage future research along this direction to further analyze SSL and cold-start impacts on AL.

\noindent\textbf{Acknowledgment}. Discussions with Giulia DeSalvo, Chih-kuan Yeh, Kihyuk Sohn, Chen Xing, and Wei Wei are gratefully acknowledged.

\clearpage
%
%
\bibliographystyle{splncs04}
\bibliography{egbib}
\end{document}